\documentclass[11pt,a4paper]{article}

\usepackage[total={13.5cm,24cm}]{geometry}

\usepackage{natbib}
\usepackage[utf8]{inputenc}
\usepackage{type1cm}
\usepackage{url}
\usepackage[small,bf]{caption} 
\usepackage{framed, ltxtable}
\usepackage[usenames]{xcolor}
\usepackage[english]{babel}
\usepackage{multirow}
\definecolor{shadecolor}{rgb}{0.95,0.85,0.15}
\usepackage{todonotes}

\usepackage[T1]{fontenc}
\usepackage[variablett]{lmodern}
\usepackage{listings} \usepackage{carolmin, lmodern}
\newcommand{\carols}{{\selectfont\cminfamily s}} 

\newcommand{\acronym}[1]{\textsc{\lowercase{#1}}}

\newcommand{\creator}[1]{\multicolumn{2}{l}{\textsl{#1}:}\\*} 
\newcommand{\opus}[3]{&#1&#2&#3\\}

\begin{document}

\title{An open diachronic corpus of historical Spanish: annotation
  criteria and automatic modernisation of spelling}

\author{Felipe Sánchez-Martínez, Isabel Martínez-Sempere,\\ Xavier Ivars-Ribes, Rafael C. Carrasco\\
Dep. Llenguatges i Sistemes Informàtics,\\
  Universitat d'Alacant, E-03071, Alacant, Spain\\ 
   \texttt{fsanchez@dlsi.ua.es}}

\date{} 

\maketitle

\begin{abstract}
  The \acronym{IMPACT}-es diachronic corpus of historical Spanish
  compiles over one hundred books ---containing approximately 8
  million words--- in addition to a complementary lexicon which links
  more than 10 thousand lemmas with attestations of the different
  variants found in the documents. This textual corpus and the
  accompanying lexicon have been released under an open license
  (Creative Commons \acronym{BY-NC-SA}) in order to permit their
  intensive exploitation in linguistic research.

  Approximately 7\% of the words in the corpus (a selection aimed at
  enhancing the coverage of the most frequent word forms) have been
  annotated with their lemma, part of speech, and modern
  equivalent. This paper describes the annotation criteria followed
  and the standards, based on the Text Encoding Initiative
  recommendations, used to the represent the texts in digital form.

  As an illustration of the possible synergies between diachronic
  textual resources and linguistic research, we describe the
  application of statistical machine translation techniques to infer
  probabilistic context-sensitive rules for the automatic
  modernisation of spelling. The automatic modernisation with this
  type of statistical methods leads to very low character error rates
  when the output is compared with the supervised modern version of
  the text.

~

\end{abstract}

\section{Introduction}
\label{sec:introduction}

Diachronic corpora are a valuable source of information with which to
understand the historical evolution of languages. Unfortunately,
diachronic collections are relatively scarce ---at least, when
compared to the overwhelming availability of resources containing
transcriptions of modern text or
speech~\citep{kocjancic09:_inter,prochazkova06:_fundam,davies10:_corpus_contem_americ_eng,francis79:_brown_corpus_manual}---,
since creating a diachronic corpus is a costly task. The transcription
of old texts must be manually reviewed because a number of features
---such as spelling variations, old fonts, deprecated characters, and
blurred text caused by stains or page transparency--- may cause the
accuracy of the automatic process of converting printed text into
computer-encoded text (commonly referred to as OCR or Optical
Character Recognition) to fall below acceptable rates. In the
particular case of Spanish, access to suitable linguistic resources
can be challenging since the most renowned on-line diachronic
resources, such as the \emph{Corpus Diacrónico del Español}
---\acronym{CORDE}~\citep{CORDE}---, and the \emph{Corpus del
  Español}~\citep{Davies2002} can be consulted via Web
interfaces\footnote{See \url{http://corpus.rae.es/cordenet.html} and
  \url{http://www.corpusdelespanol.org}, respectively.} which provide
limited querying capabilities~\citep{Davies2010}.

This paper describes the \acronym{IMPACT}-es diachronic corpus of
historical Spanish and the accompanying lexicon created by
\acronym{IMPACT},\footnote{IMproving ACcess to Text,
  \url{http://www.impact-project.eu}} a research project funded by the
European Commission under its seventh Framework Programme and focused
on the improvement of the precision of OCR and the access to
historical texts. The components developed in \acronym{IMPACT}
include:
\begin{itemize}
\item Historical language resources for nine European languages which
  have allowed significant improvements to be made ---up to a 30\%
  \citep{does13:_lexic_ocr_dutch}--- in OCR word recall rates for
  historical documents in addition to increased productivity when used
  in combination with post-correction tools.
\item A toolkit~\citep{depuydt09:_fons_verbor} with which to build
  lexical resources, tools for their deployment ---both in OCR and
  information retrieval applications---, and named entity recognition
  tools for historical documents.
\item Better OCR engines with improved technologies for image
  enhancement, binarisation, character recognition and algorithms for
  layout detection.
\item A large ground-truth data set (a collection of images mapped
  onto extremely accurate transcriptions), coupled with a
  comprehensive evaluation toolkit.
\item A framework of service and workflow layers~\citep{neudecker11}
  which enables full flexibility between all the \acronym{IMPACT}
  components.
\end{itemize}

In particular, the \acronym{IMPACT}-es corpus contains 107 Spanish
texts first printed between 1481 and 1748 and covering a
representative variety of creators and genres (prose, theatre, and
verse). The digital versions in the corpus are based on early editions
or faithful reprints of the early editions.  This content is divided
into two separate sections:
\begin{enumerate}
\item The \acronym{GT} section which compiles the 21 Spanish documents
  in the ground-truth data set created by \acronym{IMPACT}.
\item The \acronym{BVC} section which compiles 86 texts provided by
  the \emph{Biblioteca Virtual Miguel de
    Cervantes}\footnote{\url{http://www.cervantesvirtual.com}} digital
  library and which have been partially annotated as will be described
  later.
\end{enumerate}
Moreover, the corpus is complemented with a lexicon that links more
than 10 thousand lemmas ---corresponding to the most frequent word
forms---, with a representative sample of attestations in the
corpus. The full \acronym{IMPACT}-es corpus and the accompanying
lexicon are available under the Creative Commons
Attribution-NonCommercial-ShareAlike 3.0
license\footnote{\url{http://creativecommons.org/licenses/by-nc-sa/3.0/}}
and can be downloaded at the \acronym{IMPACT} Centre of Competence
Website.\footnote{\url{http://www.digitisation.eu/tools/language-resources/impact-es/}}
A compact version of the lexicon, with no quotes from the corpus, has
also been released under a dual license ---Creative Commons
Attribution-ShareAlike
3.0\footnote{\url{http://creativecommons.org/licenses/by-sa/3.0/}} and
GNU GPL v3\footnote{\url{http://www.gnu.org/licenses/gpl.html}}--- in
order to encourage its integration into free/open-source tools for the
analysis and part-of-speech tagging of historical Spanish texts
\citep{Sanchez-Marco2011} such as FreeLing \citep{Carreras2004}.

To the best of our knowledge, this is the first diachronic corpus of
historical Spanish distributed under an open license. In addition to
the aforementioned \acronym{CORDE} and \emph{Corpus del Español}, some
other collections serve specific purposes.  For example, the
\emph{Corpus de Documentos Españoles Anteriores a 1700}
---{CODEA}~\citep{borja12:_desar_codea}--- contains over 1,500
documents, written between the 12th and 17th centuries, mainly
collected from archives (for example, letters and administrative,
legal and ecclesiastic documents). The project
Website\footnote{\url{http://demos.bitext.com/codea}} provides an
interface through which to access the palaeographic and critical
editions on a single-document basis.  The \emph{Corpus Histórico del
  Español en México} ---CHEM, ~\citep{medina11:_el_mexic}--- collects
documents written between the 16th and the 20th centuries, but the
access to the collection is restricted to text
visualisation.\footnote{\url{http://www.iling.unam.mx/chem/}}

We hope that releasing the corpus under an open license will foster
its exploitation and improvement by the linguistic research community
in addition to its integration into natural language processing
applications. As an illustration of the complementarity between
linguistic digital resources and tools, we have explored the
application of statistical machine translation techniques to the
automatic modernisation of spelling. More precisely, as is described
in Section~\ref{sec:modernisation}, we have applied phrase-based
statistical machine translation methods~\citep[ch. 5]{Koehn2010SMT} in
order to infer probabilistic context-sensitive rules for the automatic
modernisation of spelling. This method could be used as the
transcription step of tools such as ToTrTaLe \citep{erjavec11:_autom}.

The following section provides an overview of the source documents in
the \acronym{IMPACT}-es corpus and describes the criteria followed in
their annotation. The markup format used to encode the documents that
integrate the corpus and the lexicon is presented in
Section~\ref{sec:tei}. In Section~\ref{sec:modernisation} we
illustrate the exploitation of the historical corpus in natural
language processing applications. Finally, some concluding remarks are
presented in Section~\ref{sec:concluding-remarks}.

\section{Content and annotation criteria}
\label{sec:content}

The \acronym{IMPACT}-es corpus was created to assist in the
improvement of OCR techniques by allowing the integration of specific
vocabularies in the digitisation process in both information retrieval
services and comprehensive evaluation frameworks. The selection of
content and the annotation criteria have therefore been oriented
towards this objective: for instance, the original spelling (even if
clearly unintentional) has been preserved by the accurate
transcription performed to create the ground-truth documents.

The \acronym{GT} section of the \acronym{IMPACT}-es corpus compiles 21
texts printed between 1543 and 1748, and contains around 6 million
(unannotated) words with a transcription accuracy (as compared to the
original text) of over 99.95\%. The creator, title and dates of the
first and source edition of the constituent documents are listed in
Appendix~\ref{sec:gt_content}.

In order to ensure 99.95\% fidelity to the source texts the
transcription followed the standard \emph{acceptance
  sampling}~\citep[part 6]{Montgomery2009} statistical procedure for
quality control: each document was processed in batches (containing
between 500 and 1200 pages) in which each page was scanned and the
automatic transcription obtained from the scanned image was manually
corrected; when the transcription of a whole batch was complete, a
sample containing about 4\% of the pages was randomly selected and
reviewed by an external quality control team; whenever the accuracy of
the digitisation was found to be below 99.95\%, the whole batch was
rejected and its processing restarted.

The \acronym{BVC} section of the corpus compiles 86 texts printed
between 1482 and 1990 
which are listed in Appendix~\ref{sec:bvc_content} together with the
dates of the first and source editions. These documents contain
approximately 2 million

words and a significant fraction of this
content 
---over 27\% of the section---, has been manually annotated with
linguistic metadata.

The metadata added to Spanish words are lemma (in modern form), part
of speech and modern equivalent. The words originating from other
languages (less than 0.1\%, and principally Latin) are labelled solely
with their language.  The part-of-speech categories and their tags,
shown in Table~\ref{tab:tagset} together with their relative
frequencies, are based on those defined by the Apertium machine
translation platform~\citep{Forcada2011} for the dictionaries in the
Spanish--Catalan language pair.

The annotation was assisted by the CoBaLT tool
\citep{kenter12:_lexic_cobal}, a Web-based editor which supports the
creation of corpus-based lexicons. The tool allows a common annotation
to be assigned to a sequence of consecutive words, and also accepts
compound lemmas and compound part-of-speech categories. The annotation
of the corpus is based on the 22nd edition of the \emph{Diccionario de
  la Lengua Española}~\citep{DRAE22} which has served as a primary
reference in determining the lemmas. For archaisms, the on-line
version of the \emph{Nuevo Tesoro Lexicográfico de la Lengua Española}
\citep{TRAE2001}, which compiles Spanish historical dictionaries
dating from 1726 (up to the 21st edition of the \emph{Diccionario de
  la Lengua Española}), was employed as a secondary reference.

\begin{table}
  \caption{Part of speech tags and their relative frequencies in the 
    annotated part of the corpus.}
  \label{tab:tagset}
  \centering
  \begin{tabular}{clr}
    \hline
    Tag & Part of speech & Frequency\\
    \hline
    \texttt{abr} &  abbreviation & 0.03\%\\
    \texttt{adj} &  adjective & 10.70\%\\
    \texttt{adv} &  adverb & 2.92\%\\
    \texttt{cnj} &  conjunction & 1.08\%\\
    \texttt{det} &  determiner & 2.64\%\\
    \texttt{ij} &  interjection & 0.19\%\\
    \texttt{n} &  noun & 29.41\%\\
    \texttt{np} &  proper noun & 7.63\% \\
    \texttt{num} & numeral & 0.39\%\\
    \texttt{pr} &  preposition & 1.59\%\\
    \texttt{prn} &  pronoun & 7.25\%\\
    \texttt{rel} & relative pronoun & 0.17\%\\
    \texttt{verb} &  verb & 36.00\%\\
    \hline
  \end{tabular}
\end{table}

The corpus is accompanied by a lexicon which links more than 10
thousand entries (simple or compound lemmas) with their attestations
in the \acronym{BVC} section of the corpus. The historical variants
are classified under lemma, part of speech and modern form (see Figure
\ref{fig:tei-dic} on page \pageref{fig:tei-dic} for an example). Each
occurrence of a variant is associated with the context in which it
appears (10 preceding and 10 trailing words) and a reference to the
document that contains it (title, author, dates of the first and
source editions). The lexicon has been generated in parallel with the
linguistic markup, and an ample coverage of lemmas and word forms was
sought. Higher priority for their inclusion in the lexicon has
therefore been given to the forms with a greater frequency.  Since
some words are ubiquitous, at most 500 occurrences per lemma have been
annotated and therefore registered in the lexicon.

The following criteria have been applied during the annotation of the
\acronym{BVC} section:
\begin{itemize}
 
\item \emph{Modern forms with compound lemmas}: word forms which
  cannot be associated with a simple lemma ---such as verbs with
  enclitic pronouns--- are marked with a compound lemma. For example
  \emph{arrepentirse} has the compound lemma
  \emph{arrepentir}+\emph{se}.

\item \emph{Word boundaries}: whenever two or more consecutive words
  in the transcription correspond to a single modern form (for example
  \emph{aun que} becomes \emph{aunque}), the word group receives a
  shared annotation.  Conversely, when one form corresponds to a
  sequence of modern forms (for example \emph{quel} becomes \emph{que
    el}), they are tagged with a compound lemma and a compound part of
  speech.
  
\item \emph{Archaisms}: when the reference dictionary~\citep{DRAE22}
  registers a word as an archaism with a modern equivalent, the modern
  form has been preferred for the lemma. For example \emph{apercibir}
  is the lemma assigned to the word form \emph{apercebir}.
  
\item \emph{Contractions}: the modern word forms \emph{al} and
  \emph{del} have been assigned a compound part-of-speech tag
  (\emph{pr}+\emph{det}), but only a single lemma.
  
\item \emph{Numbers}: all cardinal numbers share a single
  part-of-speech category (called \emph{num}). The original style
  ---alphabetic characters, Arabic or Roman figures--- has been
  preserved in the associated metadata. If they are split, as is the
  case of the historical variant \emph{diez y seis} rather than the
  modern \emph{dieciséis}, then they are handled as if they were
  different words in the sentence.
  
\item \emph{Optional diacritics}: optional diacritics used only for
  disambiguation (for example, in the words \emph{sólo}, \emph{éste},
  \emph{ése}, and \emph{aquél}) have been preserved in the associated
  metadata (lemma and modern form).
  
\item \emph{Apocopation}: the full form has been preferred as the
  lemma for words with apocopation. For example \emph{algún} is a form
  with the lemma \emph{alguno}.
  
\item \emph{Past participles}: past participles have been classified
  as verbs only if they follow an auxiliary verb (some cases of
  \emph{haber} or \emph{ser}), or they are not described as adjectives
  in the reference dictionary (e.g., \emph{dormido}).
  
\item \emph{Proper nouns}: all lemmas and modern forms, with the
  exception of proper nouns, are written in lower case letters. The
  modern form of proper nouns, when available, has been preferred in
  the annotation to the word form in the source text; for example, the
  modern form and lemma of \emph{Quixote} is \emph{Quijote}.
\end{itemize}

\section{Markup schema}
\label{sec:tei}

The \acronym{IMPACT}-es corpus is distributed as a collection of XML
files, each of which corresponds to a different work, and organised in
one folder per section. The XML standard~\citep{W3C:XML} specifies how
metadata must be inserted in a digital text in the form of tags which
serve to identify the nested \emph{elements} that make up the logical
structure of the document. The names of the elements and their
\emph{attributes} (optional features whose value can make the meaning
of the tag more specific) are defined by the annotation schema. In
this case, the markup vocabulary follows the Text Encoding Initiative
P5
guidelines.\footnote{\url{http://www.tei-c.org/release/doc/tei-p5-doc/en/html}}

The TEI vocabulary is widely and actively used in digitisation
projects in the area of humanities, example of which are: Europeana
Regia,\footnote{\url{http://www.europeanaregia.eu}} the Perseus
Digital Library,\footnote{\url{http://www.perseus.tufts.edu}} or the
British National Corpus.\footnote{\url{http://www.natcorp.ox.ac.uk}}
The TEI vocabulary defines a rich variety of elements such as
paragraphs, words, and characters, in addition to a number of optional
attributes ---such as type or language--- for each element, and only a
reduced subset has therefore been employed in this case.  Other
corpora released in the scope of the IMPACT project use a similar TEI
vocabulary \citep{erjavec12:_sloven}.

\begin{figure}
\centering
\begin{lstlisting}
<fileDesc>
  <sourceDesc>
    <bibl>
      <title>Primera parte de comedias del célebre poeta español, 
             Don Pedro Calderón de la Barca</title>  
      <author>Pedro Calderón de la Barca</author>
      <date type="first-edition">1685</date>
      <date type="source-edition">1685</date>
    </bibl>
  </sourceDesc>
  <editionStmt>  
    <!--- continued (trailing content omitted) --->
  </editionStmt> 
</fileDesc>
\end{lstlisting}
\caption{Fragment showing the bibliographical metadata encoded in the
  header of one of the documents.}
\label{fig:tei-header}
\end{figure}

Every document in the corpus is encoded in a single XML file whose
root element (with tag \emph{TEI}) contains a header (under the
element tag \emph{teiHeader}) ---with descriptive metadata of the
document (marked as \emph{fileDesc})---, and a body (\emph{body})
---with the digitised content structured in one or more divisions
(elements \emph{div})---. The descriptive metadata element includes
the bibliographical description of the source (under tag \emph{bibl})
together with the information concerning the digital edition (under
tag \emph{editionStmt}). The bibliographic descriptions, illustrated
in Figure \ref{fig:tei-header}, consist of a \emph{title} element, an
\emph{author} element, the year of the first edition (as a \emph{date}
element with a \emph{first-edition} value of the \emph{type}
attribute), and the year of the source edition (as a \emph{date}
second element with a \emph{source-edition} value of the \emph{type}
attribute).

\begin{figure}
\centering
\begin{lstlisting}
<div type="pb">
  <ab type="p">
    <w lemma="comedia" type="n">
      <choice>
        <orig>Comedia</orig>
        <reg>comedia</reg>
      </choice>
    </w>
    <w lemma="del" type="pr det">
      <choice>
        <orig>del</orig>
        <reg>del</reg>
      </choice>
    </w>
    <w lemma="príncipe" type="n">
      <choice>
        <orig>Príncipe</orig>
        <reg>príncipe</reg>
      </choice>
    </w>
    <w lemma="Inocente" type="np">
      <choice>
        <orig>Ynocente</orig>
        <reg>Inocente</reg>
      </choice>
    </w>
    <pc>.</pc>
    <w lemma="en" type="pr">
      <choice>
        <orig>En</orig>
        <reg>en</reg>
      </choice>
    </w>
    <w lemma="Madrid" type="np">
      <choice>
        <orig>Madrid</orig>
        <reg>Madrid</reg>
      </choice>
    </w>
    <w>a</w>
     <!--- continued (trailing words omitted) --->
  </ab>
</div>
\end{lstlisting}
\caption{Excerpt showing the content of one block element in the body
  of a TEI encoded document in the \acronym{BVC} section of the
  corpus.}
\label{fig:tei-texts-bvc}
\end{figure}

Figure~\ref{fig:tei-texts-bvc} illustrates the elements and attributes
employed in the annotation of the \acronym{BVC} section within a
single main division (element with a \emph{div} tag) for each
document:
\begin{itemize}
\item Anonymous blocks (an element with an \emph{ab} tag) contain one
  block of text (e.g., a paragraph or header) in the document.
  
\item Every anonymous block contains one or more words (elements with
  a \emph{w} tag).  Foreign words have a single attribute defined,
  \emph{xml:lang}, which stores the language of the word. In contrast,
  Spanish words are annotated with the following metadata:
  \begin{itemize}
  \item The lemma is stored as the value of the \emph{lemma}
    attribute.
  \item The part-of-speech category is the value of the \emph{type}
    attribute.
  \item The original and modern equivalent are provided as the content
    of \emph{orig} element tags (defined by TEI as an element with
    which to mark a piece of text as following the original) and
    \emph{reg} (defined as an element with which to mark a reading
    which has in some respect been regularised) under the
    \emph{choice} element tag.
 \end{itemize}

\item Punctuation characters (\emph{pc} tag) and other characters
  (\emph{c} tag) can appear between the words.
\end{itemize}
The subset of tags described above is similar to that used
by~\citet{Sanchez-Marco2009}.

\begin{figure}
\centering
\begin{lstlisting}
<div type="page">
  <ab type="page-number">Num. 28.</ab>
  <head>LA GRAN CENOBIA.</head>
  <head>COMEDIA FAMOSA.</head>
  <head>De Don Pedro Calderon de la Barca.</head>
  <head>PERSONAS QUE HABLAN EN ELLA.</head>
  <p>Aureliano.   
     Decio.
     Libio, Infante.
     <!--- continued (trailing text omitted) --->
  </p>
  <!--- continued (trailing paragraph and anonymous blocks omitted) --->
  <ab type="drop-capital">E</ab>
  <p>Spera sombra mia, palida imagen de mi
     <!--- continued (trailing text omitted) --->
  </p>
</div>
\end{lstlisting}
\caption{Excerpt showing one page element of a document in the \acronym{GT}
  section.}
\label{fig:tei-texts-gt}
\end{figure}

The \acronym{GT} section documents contain one division per page (see
Figure~\ref{fig:tei-texts-gt}) which accepts some sub-elements:
\begin{itemize}
\item Page number, drop capital, footnote and table-of-content entries
  are tagged as anonymous blocks with the specific values
  (\emph{page-number}, \emph{drop-capital}, \emph{footnote}, and
  \emph{TOC-entry}) of their \emph{type} attribute.
\item A heading is tagged as a \emph{head} element.
\item Paragraphs are marked with a \emph{p} tag.
\end{itemize}
Other constituents of the source document, such as figures, catchwords
---words placed at the foot of a page to anticipate the first word of
the following page--- or glosses were not digitised owing to the OCR
orientation of the corpus.

\begin{figure}
\centering
\begin{lstlisting}
<entry xml:id="id80" n="abrazarle-vblex prn">
  <form type="lemma">
    <orth>abrazarle</orth>
    <gramGrp>
      <gram type="pos">vblex</gram>
      <gram type="pos">prn</gram>
    </gramGrp>
    <lbl type="occurrences">30</lbl>
  </form>
  <form type="modern-form">
    <orth>abracele</orth>
    <form type="historical-form">
      <orth>abráçele</orth>
      <cit>
        <bibl>
          <title>Segunda Celestina</title>
          <author>Feliciano de Silva</author>
          <date type="first-edition">1536</date>
          <date type="source-edition">1536</date>
        </bibl>
        <quote><!--- text omitted---> y <oVar>abráçele</oVar> ay, <!--- text omitted---></quote>
      </cit>
    </form>
    <form type="historical-form">
      <orth>abraçele</orth>
      <cit>
        <bibl>
          <title>Viaje del Parnaso</title>
          <author>Miguel de Cervantes Saavedra</author>
          <date type="first-edition">1614</date>
          <date type="source-edition">1614</date>
        </bibl>
        <quote><!--- text omitted---> y <oVar>abraçele</oVar>, en la <!--- text omitted---></quote>
      </cit>
    </form>
  </form>
  <form type="modern-form">
    <orth>abrazandole</orth>
    <form type="historical-form">
      <orth>abraçandole</orth>
      <!--- continued (trailing citations omitted) --->
    </form>
    <form type="historical-form">
      <orth>abraçandoles</orth>
      <!--- continued (trailing citations omitted) --->
    </form>
  </form>
      <!--- continued (trailing word forms omitted) --->
</entry>
\end{lstlisting}
\caption{Example of an entry in the lexicon.}
\label{fig:tei-dic}
\end{figure}

Because of its specific nature, a different subset of elements defined
by the TEI P5 guidelines (module ``9 dictionaries'') has been employed
for the lexicon (see Figure~\ref{fig:tei-dic}). Indeed, the body of
the lexicon document consists of entries (elements with an
\emph{entry} tag) which contain:
\begin{itemize}
\item A lemma as an element with a \emph{form} name (a form is defined
  by TEI as an element which groups all the information concerning the
  written and spoken forms of one headword) and the \emph{lemma} value
  of its \emph{type} attribute.
\item A number of modern variants of the lemma, labelled as elements
  with a \emph{form} name and a \emph{modern-form} value of their
  \emph{type} attribute.
\end{itemize}

On the one hand, every \emph{form} element of a \emph{lemma} type
contains:
\begin{itemize}
\item The lemma under the \emph{orth} TEI element (which is defined as
  the orthographic form of a dictionary headword).
\item The part-of-speech category (under the \emph{gram} element that
  provides grammatical information in a \emph{gramGrp} element).
\item The number of annotated occurrences in the collection, given by
  the content of a \emph{lbl} element with an \emph{occurrences} type.
\end{itemize}

On the other hand, every \emph{form} element of a \emph{modern-form}
type contains:
\begin{itemize}
\item The orthographic variant, as the content of an \emph{orth}
  element.
\item One or more \emph{form} elements with \emph{historical-form}
  \emph{type} which contains the historical forms (orthographic
  variant) as the content of an \emph{orth} sub-element and a number
  of attestations as the content of a \emph{cit} sub-element.
\end{itemize}

Finally, every attestation contains the following information:
\begin{itemize}
\item The bibliographical information (within the \emph{bibl}
  element), i.e. the reference of the work cited.
\item A number of quotes within the \emph{quote} element in which the
  historical form is labelled with the \emph{oVar} tag.
\end{itemize}

\section{Automatic modernisation of spelling}
\label{sec:modernisation}

The lack of normalisation in the spelling of old texts poses a
challenge to information retrieval systems, since users cannot include
all the possible variants of every word in their queries. It is also
difficult for the non-expert to interpret such documents, and
modernised and critical editions are therefore often produced to
facilitate reading. Automatic modernisation seeks to minimise the cost
of creating modern editions by using rules for the updating of
spelling which can be either provided by experts in palaeography
(following a deductive approach) or can be induced from examples when
large corpora are available.

For instance, some deductive methods use phonetic matching techniques
in order to generate alternative spellings~\citep{baron08:_vard}.  In
contrast, inductive methods for language processing assume no prior
linguistic knowledge but require a large amount of data to identify a
suitable model for the transformation rules~\citep{Manning1999}.
Updating the spelling of a text can clearly be seen as a particular
case of translation or text rewriting \citep{mihov07:_effic},
although, in contrast to the translation between divergent languages,
rules to convert the spelling operate at the character level rather
than at the word level. For example, a common modernisation rule
replaces every long s (the old character ``\carols'') with a standard
s.

We have therefore explored the applicability of inductive machine
translation techniques to the task of updating the old forms in a
document. The statistical machine translation
approach~\citep{Koehn2010SMT} would appear to be a natural candidate
for this task since it can deal with the most important features of
modernisation:
\begin{itemize}
\item It is an asynchronous process, that is, it cannot be achieved on
  a letter-by-letter basis (for example, the digraph ``ph'' usually
  becomes ``f'').
\item It is non-deterministic because the replacements can differ even
  in the same context (for example, the spelling ``fijo'' must be
  sometimes preserved, whereas in other cases it must become ``hijo'').
\item It is essentially a monotonous process, in the sense that the
  transformation of a character will not show a long range dependence
  on the context.
\end{itemize}
Alternative models, such as finite-state transducers~\citep{Oncina93}
are better suited for deterministic transductions. The traditional
input-output HMMs are not prepared to handle asynchronous
transductions~\citep{Bengio94} and, although they can be extended for
that purpose, the training phase has considerable computational
costs~\citep{Bengio96}. Furthermore, since there are a number of
open-source implementations of phrase-based statistical machine
translation methods~\citep[ch. 5]{Koehn2010SMT} which can be used to
test natural language applications, we have explored the applicability
of these methods to the task of updating the spelling.

\subsection{Method overview}
The phrase-based approach translates a sentence $s$ by maximising the
probability of the result $t$. The probability $p(t|s)$ is defined in
terms of a linear combination of a number of \emph{feature functions}:
\begin{equation}
  \label{eq:loglinear}
  p(t|s)=\exp\sum_k \lambda_k h_k(s,t).
\end{equation}
Typical feature functions $h_k(s,t)$ are the logarithms of
source-to-target and target-to-source \emph{phrase}\footnote{In the
  phrase-based approach, any segment of text, even without a syntactic
  coherence, is called a phrase.} translation probabilities,
logarithms of source-to-target and target-to-source token translation
probabilities, reordering costs, the output length, the number of
phrases used in the translation, and the logarithm of the likelihood
of the output as provided by a \emph{target-language model}.

The inference process in the SMT approach consists of the following
steps:
\begin{enumerate}
\item The feature functions are estimated using a parallel corpus
  ---more precisely, the \emph{training} subset--- after token
  alignment and phrase extraction~\citep{Zens2002}.
\item The weights $\lambda_k$ are tuned in order to optimise the
  translation quality on a held-out parallel corpus ---the
  \emph{development} subset--- using the minimum error rate training
  (MERT) algorithm~\citep{Och2003}. This optimisation is traditionally
  oriented towards maximising the popular BLEU
  score~\citep{Papineni2002}, an automatic measure of translation
  quality.
\end{enumerate}
After training, the translation for a sentence $s$ is selected by
looking for the target-language sentence $t$ which maximises $p(t|s)$.

As noted previously, modernisation can be regarded as a type of
translation in which sequences of characters that already constitute
phrases and words play the traditional role of sentences: A training
sample thus consists of pairs of words (the source and the modern
forms). For phrase extraction (phrases being groups of characters
often transformed together) the characters can be aligned with a very
simple procedure: first the longest common
sub-sequence~\citep{Hirschberg1975} is used to discover which
characters are identical in both forms and can be aligned without
further character swaps; then, the remaining sub-sequences are aligned
if their alignment does not cross over the one-to-one alignments
obtained in the first step.  This simple procedure provides suitable
results and avoids the costly training phase required by standard
statistical alignment methods~\citep{och03}.

\subsection{Experiments and results}
The set of samples obtained from the lexicon was split into training,
development and testing subsets, with the sizes shown in
Table~\ref{tab:corpus}. Interestingly, the number of characters in the
target subsets (the modern words) is slightly smaller (by only 0.3\%)
than the number of characters in the source documents, even if they
contain an identical number of words.
 
\begin{table}
  \caption{Size of the training, development and testing subsets 
    (for source documents).}
\label{tab:corpus}
\centering
\begin{tabular}{lrr}
  \hline
  Subset & Words & Characters\\
  \hline
  Training & 599,126 & 3,739,262 \\ 
  Development & 5,000   & 31,416 \\
  Test  & 5,000   & 31,239\\
  \hline
\end{tabular}
\end{table}

In order to evaluate its performance, the statistical approach has
been compared with four other methods:
\begin{enumerate}
\item The text remains as it is in the source file or, in other words,
  no modernisation takes place.
\item A naive approach selects the most frequent modernised form for
  every word which is also in the training subset and preserves the
  source form in those cases in which no reference is found in the
  training subset.
\item The source text is updated using the suggested correction
  provided by a modern spell checker (Ispell version 3.3.02 with the
  Spanish dictionary version 1.11).
\item The source text is updated with a modern spell checker with the
  enhanced coverage provided by the list of words in the modern part
  of the training subset.
\end{enumerate}

\begin{table}
  \caption{Character error rate (as a percentage) 
    for the automatic modernisation of spelling.}
  \label{tab:results}
  \centering
  \begin{tabular}{lr}
    \hline
    Method & CER \\
    \hline
    No modernisation & 5.75\%\\
    Naive & 0.50\%\\
    Spellchecker & 5.91\%\\
    Spellchecker + dictionary & 5.27\%\\
    SMT & 0.21\%\\
    \hline
  \end{tabular}    
\end{table}
The accuracy of these methods is measured using the \emph{character
  error rate} (CER), defined as the minimal number of characters that
need to be modified (inserted, removed or replaced) in order to
transform the output into the target, normalised with the total number
of characters in the target.

As is shown in Table~\ref{tab:results}, an average of 5.75\% of the
characters must be modified in the source text in order to obtain the
modernised spelling. It is worth noting that the naive approach
achieves a high accuracy by simply selecting the most common spelling,
although this method fails with all unseen words that require
modernisation (0.34\% in the test set), since in these cases the input
word is copied verbatim to the output.

In contrast, the replacements suggested by the spell checker cannot be
used to modernise the spelling, since the error rate remains
comparable to the rate obtained with the unmodified text. This low
performance is not originated by an insufficient lexical coverage,
since the addition of a specific dictionary does not reduce the number
of mistakes. Corrections based on word similarity do not therefore
appear to be adequate for the modernisation of
spelling.\footnote{Ispell corrections are based on the
  Damerau–Levenshtein distance~\citep{Lowrance1975}, a measure which
  enhances the traditional edit distance~\citep{Levenshtein1965} by
  permitting the transposition of adjacent characters.}

The lowest error rate is clearly obtained with the application of the
SMT technique, which reduces the CER to less than one half of that
obtained with the naive approach. This accuracy suggests that the
statistical method identifies the essential rules needed to transform
segments of characters into their modern spelling and it can deal with
new, unseen words.

In our settings, the SMT system generated translation rules of any
length up to 8 letters, a value which showed the best compromise
between the accuracy of results and model complexity. 

About one half of the rules learnt only transferred the input to the
output verbatim but the others produced transformations like the
following: ``eio\(\rightarrow\)ejo'', ``zys\(\rightarrow\)cís'',
``euo\(\rightarrow\)evo'', ``ço\(\rightarrow\)zo'',
``çe\(\rightarrow\)ce'', ``sçe\(\rightarrow\)ce'',
``nuio\(\rightarrow\)nvio'', ``vbe\(\rightarrow\)ube'', or
``xu\(\rightarrow\)ju''. These type of rules are similar to those
proposed by experts~\citep{Sanchez-Marco2010}.

\begin{table}
  \caption{Character error rate (CER) for the SMT approach when \(N\) 
    characters are added before and after the word in old Spanish 
    whose spelling is to be modernised. The row labelled ``unaligned'' shows
    the results when the context characters are left unaligned, the one 
    labelled ``aligned'' shows the results when these characters become
    aligned (see running text).}
  \label{tab:context}
  \centering
  \begin{tabular}{c|cccccc}
    \hline
    \(N\) & 0 & 1 & 2 & 3 & 4 & 5\\
    \hline
     Unaligned & 0.21\% & 0.27\% & 0.36\% & 0.60\% & 1.41\% & 3.41\%\\
     Aligned & 0.21\% & 0.28\% & 0.32\% & 0.60\% & 2.19\% & 7.15\%\\
    \hline
  \end{tabular}    
\end{table}

A set of additional experiments has been performed in order to
identify the influence of neighbouring words on the modernisation
process, which may be especially important in languages that exhibit
external
\emph{sandhi}~\citep{matthews97:_concis_oxfor_diction_linguis} or
similar effects on spelling (such as some Indian and Celtic
languages), and may help to disambiguate those cases in which a word
has more than one possible modernisation.

Table \ref{tab:context} shows the results obtained when the words in
each sample are contextualised by adding, in addition to a special
character representing blank spaces, \(N\) characters from the
previous word as a left context and \(N\) characters from the
following one as a right context.  The left and right context
characters receive different codes to the word characters and could
remain unaligned or they could be aligned with the initial and last
character, respectively.  Consider, for example, the training sample
with \(N=2\) ``e\(_l\)l\(_l\) fijo
d\(_r\)e\(_r\)\(\rightarrow\)hijo''. The characters in this training
sample can be aligned in two different ways: one that leaves
``e\(_l\)'', ``l\(_l\)'', ``d\(_r\)'', ``e\(_r\)'' and the blank
spaces unaligned, and another that aligns all the characters in the
left context to the first letter of the modernised form (``h''), and
all the characters in the right context to the last letter of the
modernised form (``o'').

The experiments show that, in the case of Spanish, the best results
are obtained when no context is added (\(N=0\)). The analysis of the
output reveals that often, especially in the case of a large $N$, many
words never appear in the training subset with an identical context to
the test subset and, in such cases, the quality of the translation
deteriorates.

\section{Concluding remarks}
\label{sec:concluding-remarks}

The \acronym{impact}-es open diachronic corpus of historical Spanish
contains approximately 8 million words and has been released under an
open license (Creative Commons BY-NC-SA). We have described the
criteria applied for the linguistic annotation ---of nearly 7\% of the
words in the corpus--- with lemmas, parts of speech and modern
equivalents.

The corpus is divided into two sections: the \acronym{BVC} section
(from the \emph{Biblioteca Virtual Miguel de Cervantes} digital
library), which has been manually annotated, and the \acronym{GT}
section (developed by the IMPACT project), which has been digitised
with a fidelity of 99.95\% to the original. Furthermore, a lexicon has
been extracted from the \acronym{BVC} section. Every entry in the
lexicon corresponds to one lemma and part of speech, and contains a
sample of variants and their attestations in the corpus.

The application of phrase-based statistical machine translation
techniques for the modernisation of spelling has a very high accuracy
---the character error rate is below 0.25\%--- and demonstrates the
complementarity of diachronic corpora and linguistic applications. The
transformation rules which have been automatically identified are
analogous to those identified by experts. However, it is worth noting
that the statistical machine translation approach does not apply
modernisation rules based only on the probability of the translation
rules, but rather on a combination of features such as the target
language likelihood.  The best accuracy is achieved when the words are
modernised without considering the context characters from
neighbouring words. This observation is consistent with the fact that
99.11\% of the words in the training subset have only one possible
modernisation.

This automatic modernisation achieves sufficient accuracy for the
development of useful tools to assist in the production of modernised
and critical editions, or to alleviate the difficulties that searching
and retrieving texts with multiple historical variants creates.

\appendix
\newpage

\section{Content: the GT section}
\label{sec:gt_content} 

\begin{longtable}[p]
  {p{0\textwidth}p{0.6\textwidth}cc}

  \label{tab:groundtruth} \\
  \hline
    & \multirow{2}{*}{Author: Title} & First & Source\\
    &               & edition & edition\\
  \hline
  \endfirsthead
  \hline
    & \multirow{2}{*}{Author: Title} & First & Source\\
    &               & edition & edition\\
  \hline
  \endhead
  \hline
  \multicolumn{4}{r}{Continued on next page \(\ldots\)}
  \endfoot
  \hline
  \endlastfoot
  \creator{Anonymous}
        \opus{Vida de Lazarillo de Tormes}{1554}{1652}
  \creator{Francisco de Quevedo}
             \opus{El Parnasso español}{1648}{1648}
  \creator{Garcilaso de la Vega}
             
             \opus{Obras de Garcilasso de la Vega con las anotaciones por el Mtro. Francisco Sánchez Brocense}{1574}{1612}
  \creator{Inca Garcilaso de la Vega} 
              \opus{Commentarios reales}{1609}{1609}
  \creator{Jorge Juan} 
               \opus{Observaciones astronomicas y phisicas hechas de orden de S. M. en los Reynos del Peru}{1748}{1748}
  \creator{Juan Boscán} 
              
              \opus{Las obras de Boscán y algunas de Garcilasso de la Vega repartidas en cuatro libros}{1543}{1543} 
  \creator{Lope de Vega} 
              \opus{Las comedias del famoso poeta Lope de Vega}{1604}{1604}
  \creator{Luis de Góngora} 
              \opus{El Polifemo de Don Luis de Góngora with comments by Don García de Salzedo}{1629}{1629}
  \creator{Mateo Alemán} 
               \opus{Vida y hechos del pícaro Guzmán de Alfarache}{1599}{1681}
  \creator{Miguel de Cervantes Saavedra} 
                \opus{El ingenioso hidalgo Don Quixote de la Mancha}{1605}{1605}
  \creator{Pedro Calderón de la Barca}
             \opus{Primera parte de comedias del célebre poeta español, Don Pedro Calderón de la Barca}{1685}{1685}
  \creator{Real Academia Española de la Lengua} 
              \opus{Diccionario de la lengua castellana [...] Tomo primero. Que contiene las letras A, B}{1726}{1726}
              \opus{Diccionario de la lengua castellana [...] Tomo segundo. Que contiene la letra C}{1729}{1729}
              \opus{Diccionario de la lengua castellana [...] Tomo tercero. Que contiene las letras D, E, F}{1732}{1732}
              \opus{Diccionario de la lengua castellana [...] Tomo quarto. Que contiene las letras G, H, I, J, K, L, M, N}{1734}{1734}
              \opus{Diccionario de la lengua castellana [...] Tomo quinto. Que contiene las letras O, P, Q, R}{1737}{1737}
              \opus{Diccionario de la lengua castellana [...] Tomo sexto. Que contiene las letras S, T, V, X, Y, Z}{1739}{1739}
  \creator{Ruy López de Sigura}
              \opus{Libro de la invención liberal y arte del juego del Axedrez}{1561}{1561}
  \creator{San Juan de la Cruz} 
             \opus{Obras del venerable y mistico Dotor F. Joan de la Cruz}{1629}{1629}
  \creator{Santa Teresa de Jesús} 
               \opus{Los libros de la Madre Teresa de Jesús}{1588}{1588}
  \creator{Sor Juana Inés de la Cruz} 
             \opus{Carta athenagorica}{1690}{1690}
\end{longtable}
\newpage

\section{Content: the BVC section}
\label{sec:bvc_content}

\begin{longtable}[p]
  {p{0\textwidth}p{0.6\textwidth}cc}

  \small
  \label{tab:cervantes} \\
  \hline
    & \multirow{2}{*}{Author: Title} & First & Source\\
    &               & edition & edition\\
  \hline
  \endfirsthead
  \hline
    & \multirow{2}{*}{Author: Title} & First & Source\\
    &               & edition & edition\\
  \hline
  \endhead
  \hline
  \multicolumn{4}{r}{Continued on next page \(\ldots\)}
  \endfoot
  \hline
  \endlastfoot
  \creator{Baltasar Gracián} 
	\opus{Oráculo manual y arte de la prudencia}{1647}{1647}
  \creator{Beato Juan de Ávila} 
	\opus{Epistolario espiritual}{1578}{1962}
  \creator{Cristóbal de Castillejo} 
	\opus{Dialogo de mujeres}{1544}{1544}
	\opus{Obras morales y de devoción}{1542}{1958}
  \creator{Diego Sánchez de Badajoz} 
	\opus{Farsa de Abraham}{1554}{1554}
	\opus{Farsa de la muerte}{1554}{1554}
	\opus{Farsa racional del libre alvedrío}{1554}{1554}
  \creator{Feliciano de Silva} 
	\opus{Segunda Celestina}{1536}{1536}
  \creator{Fernando Rojas} 
	\opus{La Celestina}{1499--1502}{1499, 1514}
  \creator{Fernán Pérez de Oliva} 
	\opus{Dialogo de la dignidad del hombre}{1585}{1586}
  \creator{Francisco de la Torre} 
	\opus{Poesías}{Various}{1969}
  \creator{Francisco Delicado} 
	\opus{La Lozana Andaluza}{1528}{1528}
  \creator{Gabriel Lobo Lasso de la Vega} 
	\opus{Tragedia de la honra de Dido restaurada}{1587}{1587}
  \creator{Guillén de Castro} 
	\opus{Las Mocedades del Cid}{1605--1615}{1618}
  \creator{Íñigo de Mendoza} 
	\opus{Coplas de Vita Christi Frayy}{1482}{1482}
  \creator{Juan Boscán} 
	\opus{Obra completa}{Various}{1917}
  \creator{Juan Cortés de Tolosa} 
	\opus{El desgraciado}{1617}{1620}
	\opus{El nacimiento de la verdad}{1617}{1620}
	\opus{La Comadre}{1617}{1620}
	\opus{Novela del licenciado periquín}{1617}{1620}
	\opus{Novela de un miserable llamado Gonzalo}{1617}{1620}
  \creator{Juan de Encina} 
	\opus{Égloga representada en la noche postrera de Carnal}{1496}{1496}
	\opus{Aucto del repelón}{1509}{1509}
	\opus{Égloga de Cristino y Febea}{1509}{1509}
	\opus{Égloga de Fileno, Zambardo y Cardonio}{1509}{1509}
	\opus{Égloga de las grandes lluvias}{1507}{1507}
	\opus{Égloga de Mingo, Gil y Pascuala}{1496}{1496}
	\opus{Égloga de Plácida y Vitoriano}{1513}{1962}
	\opus{Representación sobre el poder del amor}{1507}{1507}
  \creator{Juan de Mena} 
	\opus{Laberinto de Fortuna}{1481}{1505}
  \creator{Juan Ruiz de Alarcón y Mendoza} 
	\opus{El antichristo}{1634}{1990}
	\opus{El desdichado en fingir}{1628}{1990}
	\opus{El dueño de las estrellas}{1634}{1990}
	\opus{El tejedor de Sevilla}{1634}{1990}
	\opus{Examen de maridos}{1634}{1990}
	\opus{Ganar amigos}{1634}{1990}
	\opus{La amistad castigada}{1634}{1990}
	\opus{La crueldad por el honor}{1634}{1990}
	\opus{La cueva de Salamanca}{1628}{1990}
	\opus{La industria y la suerte}{1628}{1990}
	\opus{La manganilla de Melilla}{1634}{1990}
	\opus{La prueba de las promesas}{1634}{1990}
	\opus{Los empeños de un engaño}{1634}{1990}
	\opus{Los pechos privilegiados}{1634}{1990}
	\opus{Mudarse por mejorarse}{1628}{1990}
	\opus{Todo es ventura}{1628}{1990}
  \creator{Lope de Vega} 
	\opus{Comedia del Príncipe Ynocente}{1590}{1762}
  \creator{Luis Vélez de Guevara} 
	\opus{La serrana de la Vera}{1613}{1916}
  \creator{Miguel de Cervantes Saavedra} 
	\opus{Comedia del çerco de Numancia}{1615}{1615}
	\opus{Comedia famosa de la casa de los zelos y seluas de Ardenia}{1615}{1615}
	\opus{Comedia famosa del gallardo español}{1615}{1615}
	\opus{Comedia famosa del laberinto de amor}{1615}{1615}
	\opus{Comedia famosa de los baños de Argel}{1615}{1615}
	\opus{Comedia famosa de Pedro de Vrdemalas}{1615}{1615}
	\opus{Comedia famosa intitvlada el rvfian Dichoso}{1615}{1615}
	\opus{Comedia famosa intitvlada la gran svltana doña Catalina de Ouiedo}{1615}{1615}
	\opus{Comedia llamada Trato de Argel}{1615}{1615}
	\opus{Don Quijote de la Mancha (1ª parte)}{1605}{1605}
	\opus{Don Quijote de la Mancha (2ª parte)}{1615}{1615}
	\opus{Entremes de la cueua de Salamanca}{1615}{1615}
	\opus{Entremes de la eleccion de los alcaldes de Daganço}{1615}{1615}
	\opus{Entremes de la guarda cuydadosa}{1615}{1615}
	\opus{Entremes del juez de los diuorcios}{1615}{1615}
	\opus{Entremes del retablo de las marauillas}{1615}{1615}
	\opus{Entremes del rufian viudo, llamado Trampagos}{1615}{1615}
	\opus{Entremes del viejo zeloso}{1615}{1615}
	\opus{Entremes del vizcayno fingido}{1615}{1615}
	\opus{La entretenida}{1615}{1615}
	\opus{La Española inglessa}{1613}{1613/1614}
	\opus{La Galatea}{1585}{1585}
	\opus{Novela de la Fuerça de la sangre}{1613}{1613/1614}
	\opus{Novela de la Gitanilla}{1613}{1613/1614}
	\opus{Novela de la Illustre Fregona}{1613}{1613/1614}
	\opus{Novela del amante liberal}{1613}{1613/1614}
	\opus{Novela de las dos Donzellas}{1613}{1613/1614}
	\opus{Novela de la Señora Cornelia}{1613}{1613/1614}
	\opus{Novela del Casamiento engañoso}{1613}{1613/1614}
	\opus{Novela del Licenciado Vidriera}{1613}{1613/1614}
	\opus{Novela del Zeloso estremeño}{1613}{1788}
	\opus{Novela de Rinconete y Cortadillo}{1613}{1788}
	\opus{Novelas exemplares}{1613}{1613/1614}
	\opus{Novela y coloquio que passó entre Cipion y Bergança, perros del hospital de la Resureccion}{1613}{1613/1614}
	\opus{Ocho comedias y ocho entremeses nuevos}{1615}{1615}
	\opus{Persiles y Sigismunda}{1617}{1617}
	\opus{Poesías sueltas}{1615}{1615}
	\opus{Viaje del Parnaso}{1614}{1614} 
\end{longtable}

\section*{Acknowledgements}
  Work funded by the European Commission under the Seventh Framework
  Programme (FP7) through the IMPACT (IMproving ACcess to Text)
  project. We thank Mikel L. Forcada for his fruitful suggestions.

\bibliographystyle{spbasic}

\bibliography{opencorpus} 

\end{document}